\pdfoutput=1
\documentclass[11pt]{article}

\usepackage[preprint]{acl}

\usepackage{times}
\usepackage{latexsym}

\usepackage[T1]{fontenc}
\usepackage[utf8]{inputenc}

\usepackage{microtype}

\usepackage{inconsolata}

\usepackage{graphicx}
\usepackage{amsmath}
\usepackage{algorithm}
\usepackage{algorithmic}
\usepackage{xcolor}
\usepackage{listings}
\usepackage{booktabs}
\usepackage{latexsym}

\usepackage{authblk}
\usepackage{booktabs}
\usepackage{multirow}
\usepackage{mathtools}
\usepackage{subfigure}
\usepackage{balance}
\usepackage{color}
\usepackage{xcolor}
\usepackage{enumitem}
\usepackage{xspace}
\usepackage{setspace}
\usepackage{amssymb}
\usepackage{longtable}
\usepackage[symbol]{footmisc}
\usepackage{tcolorbox}

\begin{document}

\title{Schema as Parameterized Tools for Universal Information Extraction}

\author{
  Sheng Liang, Yongyue Zhang, Yaxiong Wu, Ruiming Tang, Yong Liu \\
  Huawei Noah’s Ark Lab\\
  \texttt{liangsheng16@huawei.com} \\
}

\maketitle

\begin{abstract}

Universal information extraction (UIE) primarily employs an extractive generation approach with large language models (LLMs), typically outputting structured information based on predefined schemas such as JSON or tables.
UIE suffers from a lack of adaptability when selecting between predefined schemas and on-the-fly schema generation within the in-context learning paradigm, especially when there are numerous schemas to choose from.
In this paper, we propose a unified adaptive text-to-structure generation framework, called Schema as Parameterized Tools (SPT), which reimagines the tool-calling capability of LLMs by treating predefined schemas as parameterized tools for tool selection and parameter filling.
Specifically, our SPT method can be applied to unify closed, open, and on-demand IE tasks by adopting Schema Retrieval by fetching the relevant schemas from a predefined pool, Schema Filling by extracting information and filling slots as with tool parameters, or Schema Generation by synthesizing new schemas with uncovered cases.
Experiments show that the SPT method can handle four distinct IE tasks adaptively, delivering robust schema retrieval and selection performance. SPT also achieves comparable extraction performance to LoRA baselines and current leading UIE systems with significantly fewer trainable parameters. 
\end{abstract}

\section{Introduction}

Universal information extraction (UIE) primarily employs a task-agnostic extractive generation approach designed to handle various information extraction (IE) tasks in a unified and adaptable manner with large language models (LLMs).
The UIE systems usually operate across three distinct paradigms: (1) Closed-schema IE for structured templates~\cite{yadav-bethard-2018-survey, zhong-chen-2021-frustratingly, han-etal-2020-data}, (2) Open-schema IE to discover novel entities/relationships~\cite{Banko2007OpenIE, fader-etal-2011-identifying, stanovsky-etal-2018-supervised}, and (3) On-demand IE where extraction targets are dynamically specified through natural language instructions~\cite{jiao-etal-2023-instruct}. 
UIE has demonstrated superior \textit{schema adaptability} compared to traditional IE systems~\cite{li2023evaluating} that are tailored for specific tasks such as named entity recognition (NER), relation extraction (RE), and event extraction (EE). 
UIE can handle predefined schemas (structured formats) while also adapting to evolving schemas or generating new ones.

\begin{figure}[t]
  \includegraphics[width=0.96\columnwidth]{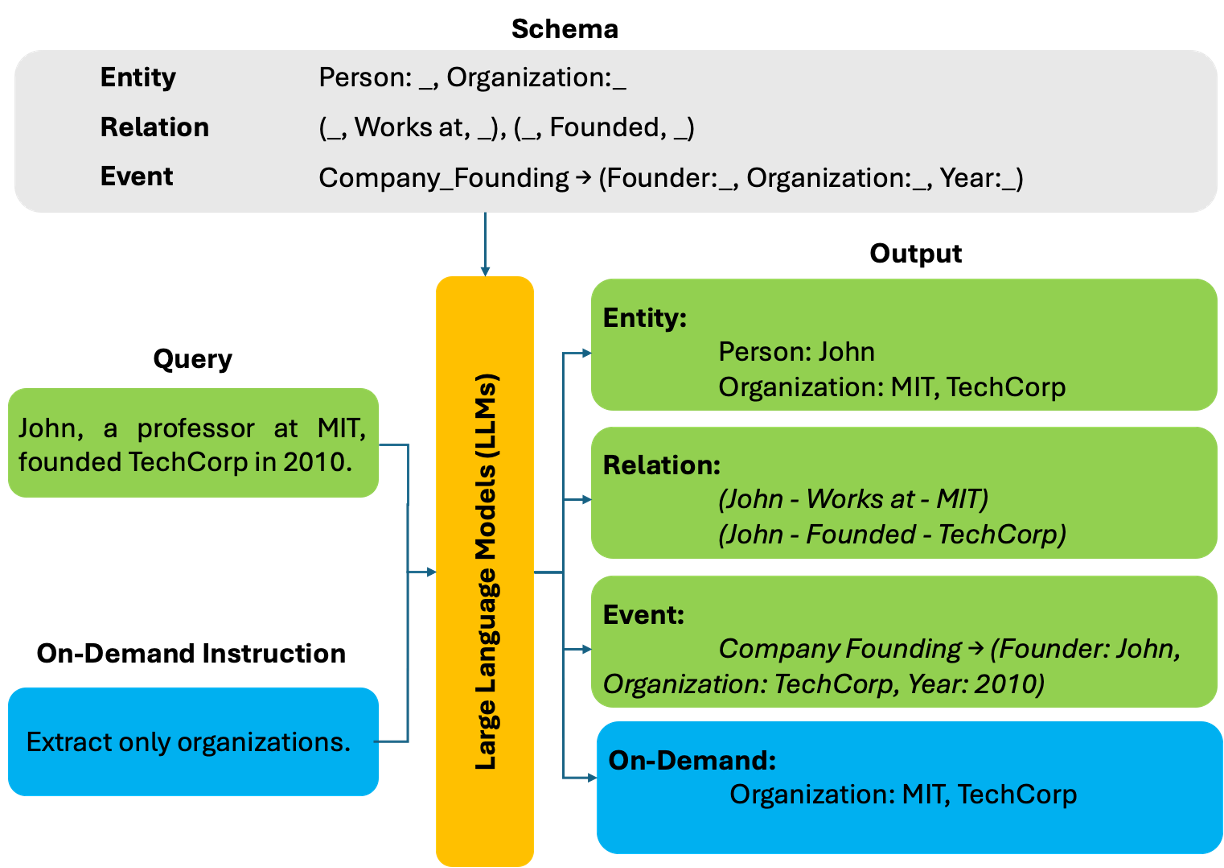}
  \caption{An overview of UIE.}
  \label{fig:task}
  \vspace{-1\baselineskip}
\end{figure}

UIE typically achieves schema adaptability by either fine-tuning large pre-trained models (LLMs) with predefined schema demonstration data or adopting the in-context learning paradigm.
However, the former paradigm restricts the extraction capability of large models to a predefined set of schemas, while the latter is constrained by the limited context length, allowing only a few demonstration shots (such as through retrieval-augmented generation (RAG)), which leads to suboptimal understanding of the extraction schemas.
In addition, UIE usually struggles with complex and unclear IE instructions~\cite{pang-etal-2023-guideline, xu2024large, sainz2024gollie}, as schema-free generation leads to unstable outputs and compromises consistency for downstream data governance, such as building a database or knowledge graph.
To the best of our knowledge, no IE system can dynamically select from numerous predefined schemas and generate schemas on the fly while ensuring governance.

Recently, tool calling has become a popular paradigm for enhancing the capabilities of LLMs, assisting in the completion of complex tasks by invoking external tools.
In particular, tool calling consists of three complementary and compatible stages: Tool Retrieval, which recalls tools relevant to the current query; Tool Creation, which generates new tools; and Tool Execution, which executes and utilizes tools to complete tasks.
For instance, ToolKenGPT~\cite{hao2023toolkengpt}treats each tool as a token (``toolken'') with a learned embedding, enabling tool calls like regular word tokens, and once triggered, prompts the LLM to complete its execution arguments.
ToolKenGPT combines the benefits of both supervised fine-tuning and in-context learning while addressing the limitations of the restricted predefined tools and limited context length.
Handling universal information extraction dynamically can be transformed into a tool-calling paradigm, offering the flexibility to integrate an arbitrary number of schemas by expanding the schema set on the fly.

In this paper, we propose a unified adaptive text-to-structure generation framework, called Schema as Parameterized Tools (SPT), which reimagines UIE through the LLM's tool-calling capacity~\cite{schick2023toolformer}, where predefined schemas act as parameterized tools, and extraction mirrors the capabilities of tool selection and parameter filling.
Additionally, inspired by the token generation style tool calling paradigm~\cite{hao2023toolkengpt}, we embed schemas as tokens to enable efficient retrieval and generation with fewer hallucinations. 
Our key insight is that the parameterized tool-calling mechanism enabling LLMs to dynamically retrieve, select, and invoke tools can be applied to unify closed, open, and on-demand IE tasks.
When processing a query, like a tool retrieval, \textbf{Schema Retrieval} fetches the top-k relevant schemas from a predefined pool. 
For uncovered cases, the LLM triggers \textbf{Schema Generation} to synthesize new schemas, effectively creating new ``tools.'' 
The LLM then performs \textbf{Schema Infilling} by extracting information and filling slots as with tool parameters. 
Our approach demonstrates strong performance across four tasks, such as Named Entity Recognition (NER), Event Extraction (EE), Relation Extraction (RE), and On-demand IE (ODIE), on four well-known IE datasets. 

The main contributions of this paper are:
\begin{itemize}
    \item We propose a unified and effective UIE framework, Schema as Parameterized Tools (SPT), which mirrors schemas as callable tools to handle all IE paradigms through a single adaptive architecture.
    \item We treat schemas as trainable token embeddings and perform efficient fine-tuning to learn the capabilities for schema retrieval and infilling.
    \item We perform extensive experiments on four well-known IE datasets that show the SPT method can handle four distinct IE tasks adaptively, delivering robust schema retrieval and selection performance. 
\end{itemize}

\section{Related Work}

\paragraph{LLM-based UIE: Flexibility at a Cost} 
In the pre-LLM era, information extraction systems focused on tasks like Named Entity Recognition (NER)~\cite{TjongKimSang2003IntroductionTT}, Relation Extraction (RE)~\cite{mintz-etal-2009-distant}, and Event Extraction (EE)~\cite{ahn-2006-stages}. 
These methods usually rely on sequence-tagging architectures~\cite{mcclosky-etal-2011-event-extraction,li-etal-2013-joint,nguyen-etal-2016-joint-event}, while achieving strong performance, they require laborious schema-specific word-level annotation and suffered catastrophic performance drops when the schemas evolved. 
With the rise of large language models (LLMs), IE has seen significant advances, especially in tasks that require greater flexibility and adaptation, by either fine-tuning LLMs with predefined schema or adopting the in-context learning paradigm.

The fine-tuning approaches, like UIE~\cite{lu-etal-2022-unified}, YAYI-UIE~\cite{xiao2023yayi}, KnowCoder~\cite{li-etal-2024-knowcoder}, and IEPile~\cite{gui-etal-2024-iepile}, fine-tune LLMs on large-scale IE corpus with instructions, achieving generalization capabilities on various IE scenarios. ADELIE~\cite{qi-etal-2024-adelie} further involves reinforcement learning to improve extraction quality. Although these methods uniformly model different information extraction tasks, their heavy architectures suffer from computational efficiency and lack a flexible framework to tackle extraction with unclear or no instructions.

The in-context learning paradigm allows for a few-shot approach, where schema demonstrations are provided in the prompt to instruct how to use the schemas.
In particular, the retrieval-augmented generation (RAG) approaches~\cite{efeoglu2024retrieval,guo2023retrieval,shiri2024decompose,gao2023retrieval} enhance the ability of LLMs to retrieve relevant few-shot examples from a large pool of query-schema-result pairs. 
By searching for semantically similar queries to the input, the system can leverage these retrieved examples in a few-shot setting to improve extraction accuracy. 
However, they inherit the limitations of their example pools and do not scale well to unseen schema types. 
Moreover, none dynamically select between predefined schemas and on-demand schema generation — a capability our work introduces through tool-calling mechanisms.

\paragraph{Tool Calling: A Missing Link for Adaptive IE} 
The concept of tool-calling with LLMs has gained traction, where LLMs invoke external tools (or schemas) to assist with tasks. These architectures introduce a novel way to handle information extraction dynamically.

\textit{Tool Retrieval} acts as the pre-stage of tool calling, utilizing dense retrieval models to recall the most relevant tools from the rich tool library based on semantic similarity to the query~\cite{zheng-etal-2024-toolrerank, xu-etal-2024-enhancing-tool}. 
This preliminary screening reduces the difficulty of tool selection for LLM, analogous to our schema retrieval phase but limited to predefined tools.

\textit{Tool Creation}~\cite{cai2024large, qian-etal-2024-toolink, yuan2024craft} aims to call tools that are not predefined, by generating new tools for unseen tasks. 
While focusing on API generation rather than structured data extraction, this approach inspires our schema generation process. 
Tool creation mirrors the need for adaptive schema generation in dynamic environments, providing a robust solution when predefined schemas are insufficient.

\textit{Tool Execution}~\cite{schick2023toolformer,hao2023toolkengpt,liu2025toolace} is a key step in tool calling, as it executes and utilizes tools to complete tasks. 
Specifically, parameter filling for predefined tools in tool execution closely aligns with the information extraction task based on predefined schemas. 
The accuracy of tool parameter filling determines the effectiveness of tool execution. 
Unlike tool calling, the information extraction task is considered complete once the parameter filling is done, without requiring the full execution result of tools.

Tool calling is an emerging paradigm where LLMs invoke external tools to assist in various tasks. 
Frameworks like ToolFormer~\cite{schick2023toolformer}, ToolKenGPT~\cite{hao2023toolkengpt}, and ToolACE~\cite{liu2025toolace} train an LLM to call external tools, demonstrate LLMs’ ability to invoke tools via parameter infilling, mirroring our schema infilling mechanism. 
ToolKenPlus~\cite{yakovlev-etal-2024-toolken} further enables LLMs to dynamically select tools with a reject option, the two-stage framework allows handling evolving tool APIs.
Our key innovation lies in reconceptualizing schemas as tools, bridging the tool-calling paradigm with IE needs. 
We introduce schema-token alignment for efficient retrieval and extraction, maintaining data governance compliance through adaptive schema selection and generation.

\paragraph{PEFT: Parameter-Efficient Fine-Tuning of LLMs}

PEFT (Parameter-Efficient Fine-Tuning)~\cite{xu2023parameter,ding2023parameter,han2024parameter} optimizes large language models (LLMs) by updating only a small subset of parameters, enabling efficient adaptation to new tasks with minimal computational resources, which is suitable for our IE schema token embedding method.
PEFT (Parameter-Efficient Fine-Tuning) methods primarily include LoRA (Low-Rank Adaptation)~\cite{hu2021lora}, which adjusts specific weight matrices through low-rank decomposition to reduce parameter updates and computational cost; Adapter Layers~\cite{pfeiffer2020adapterhub}, which insert small trainable adaptation layers between pretrained model layers to enable task adaptation without major parameter modifications; Prefix-Tuning~\cite{li2021prefix}, which prepends trainable prompt embeddings to input data, allowing the model to adjust its behavior during inference without altering core parameters; Prompt-Tuning~\cite{lester2021power}, which optimizes a set of trainable soft prompts (embedding vectors) to guide pretrained models in task execution, particularly for large language models (LLMs); BitFit~\cite{zaken2021bitfit}, which fine-tunes only bias terms in Transformer layers for highly efficient parameter tuning.
To the best of our knowledge, we are the first to explore efficient tuning methods for predicting schemas as tokens for schema learning of massive schemas.

\section{Methodology}
\label{sec:methodology}

\begin{figure*}[!ht]
\centering
\includegraphics[width=0.94\textwidth]{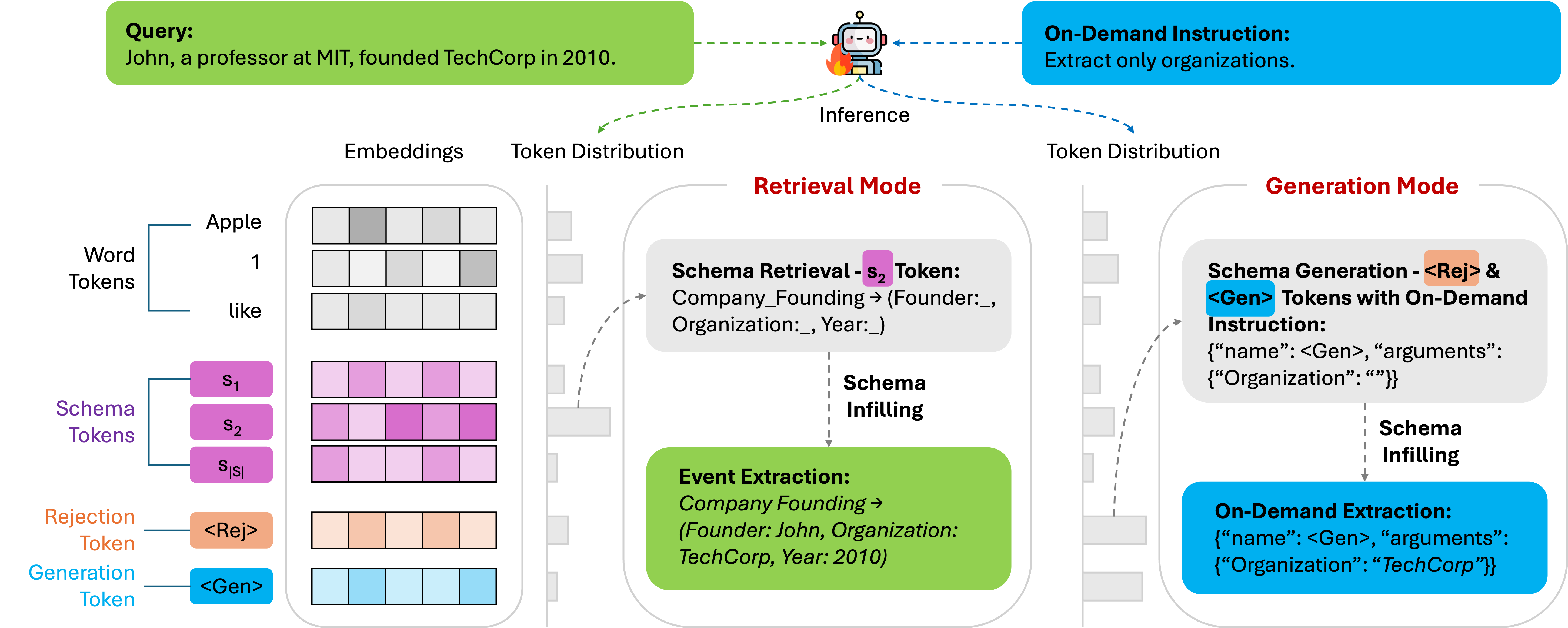}
\caption{Overview of our proposed Schema as Parameterized Tools (SPT) framework. Schema-token embeddings are appended to the LLM's head as regular word tokens. The inference procedure shows a dual-mode extraction with Retrieval and Generation Modes, consisting of Schema Retrieval, Schema Generation, and Schema Infilling.}
\label{fig:arch}
\vspace{-1\baselineskip}
\end{figure*}

In this section, we present Schema as Parameterized Tools (SPT), which enable LLMs to learn and use massive schemas for universal information extraction (UIE) with flexibility and schema adaptability.
We begin by introducing our notations and formulating the problem of universal information extraction (UIE) via tool use with LLMs. 
Typically, the next token probability distribution of the LLM is $P(X)=\sum_{i=1}^{|X|} P(x_{i}|x_{<i})$, where $X=(x_{1},x_{2},...,x_{|X|})$ is a sequence of word tokens, each word token $x_{i} \in V$ is from the vocabulary $V$ of the LLM, and $x_{<i}$ denotes the partial word token sequence before $i$-th step.
Given a set of IE schemas (schema-tokens) $S=\{s_{1},s_{2},...,s_{|s|}\}$, our goal is to enable LLMs to call a subset of these IE schemas for completing the universal information extraction tasks.
Each schema-token is parameterized as a token embedding vector, we denote a set of schema token embeddings as a matrix, i.e. $W_{S} \in \mathbb{R}^{|S| \times d}$.
In addition, we also define two helper tokens, $\texttt{<Rej>}$ and $\texttt{<Gen>}$, to determine whether a suitable schema exists in S and to guide the generation of a new schema, respectively. To perform schema-based information extraction during generation, the LLM first retrieves/generates a schema and then infills the arguments.

\subsection{Framework Overview}
The core idea of Schema as Parameterized Tools (SPT) is explicitly formulating IE schemas as tokens (termed schema-tokens), drawing inspiration from~ Toolken~\cite{hao2023toolkengpt} and Toolken+~\cite{yakovlev-etal-2024-toolken}.
As illustrated in Fig.~\ref{fig:arch}, our framework integrates three key components: schema retrieval, generation, and infilling, enabling adaptive and universal IE capabilities.
Assuming trained schema token embeddings (detailed in Section 3.4), the framework demonstrates inference operation through two primary execution paradigms:

\textbf{Retrieval Mode}:
During inference, this LLM iteratively fetches relevant schemas from a predefined schema pool (i.e., \textbf{Schema Retrieval}), ensuring efficient alignment between input queries and applicable schema structures. If no predefined schema matches, the rejection token $\texttt{<Rej>}$ is generated.

\textbf{Generation Mode}:
Triggered by the $\texttt{<Rej>}$ token in uncovered cases, the LLM synthesizes customized schemas for the query (i.e., \textbf{Schema Generation}). This mode extends coverage to unseen schema configurations while maintaining structural compatibility.

Both modes conclude with \textbf{Schema Infilling}, where the LLM continues inference, systematically infills required arguments associated with retrieved/generated schemas through iterative decoding, maintaining structural consistency throughout the generation process.

This dual-mode architecture combines the efficiency of retrieval systems with the flexibility of generative approaches, establishing a robust framework for both conventional and emerging IE scenarios.
In particular, our SPT framework adapts the tool-calling paradigm for adaptive IE through three key innovations:
(1) Schema-token Embeddings (Section 3.2): Treat predefined schemas as tokens in the extended LLM vocabulary.
(2) Dual-Mode Execution (Section 3.3): Dynamic switching between predefined schema retrieval and on-the-fly schema generation via learned \texttt{<Rej>} and \texttt{<Gen>} tokens.
(3) Compositional Training (Section 3.4): Joint optimization of schema retrieval, generation in a unified token space.

\subsection{Schema-Token Embeddings}
Inspired by \citet{hao2023toolkengpt,yakovlev-etal-2024-toolken}, but tailored for IE, we extend the LLM's vocabulary with schema tokens $S=\{s_1,...s_{|S|}\}$ for predefined schemas, and two helper tokens \texttt{<Rej>} \texttt{<Gen>} for dual-mode execution. 
The embedding matrix becomes 
$$W = [W_V | W_S^*] \in \mathbb{R}^{(|V|+|S|+2) \times d}$$
where $W_V\in\mathbb{R}^{|V|\times d}$ is the original embedding matrix, $W_S^* = [W_S|w_{\texttt{<Rej>}}, w_{\texttt{<Gen>}}] \in \mathbb{R}^{|S|+2 \times d}$ is the trainable extended schema embeddings and the two helper tokens, and $d$ is the embedding dimension.
Therefore, the next token probability distribution of LLM is
$$P(x_i|x_{<i}) = softmax(W\cdot h_{i-1})$$
Recent work~\cite{wang2024toolgen} has demonstrated that this inference process does not alter the reasoning capabilities of the LLM. 
The LLM will only switch to SPT when provided with a task-specific prompt. 
We optimize only the new embedding parameters via
$$\min_{W_S^*}\sum\nolimits_{X\in \mathcal{D}} \sum\nolimits_{i=1}^{|X|} \log P(x_i | x_{<i})$$
where $D$ is the dataset and $X$ represents the query sequence. 

\subsection{Dual-Mode Execution}

To handle uncovered schemas and enable dynamic schema adaptation during inference, the model predicts the next token based on the current state. 
When the rejection token \texttt{<Rej>} is predicted, it signals that no predefined schema matches the query and triggers schema generation. 
We further introduce a pseudo schema token \texttt{<Gen>} for new schema generation, to handle uncovered schemas and enable dynamic schema adaptation. 
By introducing \texttt{<Rej>} and \texttt{<Gen>} tokens, the UIE inference process can act in a dual-mode execution: \textit{Retrieval Mode} and \textit{Generation Mode}.

If a token from $V\cup S$ is predicted, the sequence continues as expected (either as part of the CoT or by infilling arguments during tool calling). 
The dual-mode extraction process follows:
\paragraph{Retrieval Mode}
In this mode, the LLM predicts the next token
$$x_i = \begin{cases}
\arg\max_{x_i} P(x_i|x_{<i}))&\text{if } x_i \in \mathcal{V\cup S}\\
\texttt{<Rej>} & \text{otherwise}
\end{cases}$$
\paragraph{Generation Mode}
If the \texttt{<Rej>} token is predicted, the LLM switches to generation mode, which generates argument roles for the pseudo schema \texttt{<Gen>}
$$\text{schema}=\text{LLM}(X,\texttt{<Gen>})$$

Schema Infilling continues both modes to infill the arguments, as in the tool-calling process.
\subsection{Compositional Training Strategy}
To jointly optimize the schema retrieval, generation, and infilling in a unified token space, we introduce a compositional training strategy.
In particular, the training process is divided into:
\paragraph{Phase 1} We first optimize $W_S$ on data where all samples involve closed-schema extraction. Ensuring the tokens $S$ well align with the actual schemas. 
\paragraph{Phase 2} After freezing $W_S$, we train $w_{<Rej>}$ and $w_{<Gen>}$ as continuous prompt vector for on-the-fly schema generation and extraction. This phase focuses on allowing the model to dynamically create new schemas when necessary.
\paragraph{Phase 3} We jointly fine-tune $W^*_S$ with a reduced learning rate (by a factor of 10) to allow the model to optimize these components together, ensuring the effective use of predefined and generated schemas in adaptive extraction tasks. 

This adaptive training strategy enables the model to flexibly perform information extraction with both predefined and dynamically generated schemas, offering robust adaptability to various extraction tasks.

\section{Experiment}

We describe the experimental setups to evaluate the effectiveness of our proposed SPT approach for universal information extraction (UIE) in comparison to existing approaches.

\subsection{Datasets}
We perform extensive experiments on four distinct datasets tailored to different IE tasks: CrudeOilNews~\cite{lee-etal-2022-crudeoilnews} for Event Extraction (EE), SciERC~\cite{luan-etal-2018-multi} for Relation Extraction (RE), AnatEM~\cite{DBLP:journals/bioinformatics/PyysaloA14} for Name Entity Recognition (NER) and ODIE~\cite{jiao-etal-2023-instruct} for on-demand IE.
\begin{itemize}
\item \textbf{CrudeOilNews}~\cite{lee-etal-2022-crudeoilnews}: Oil market event extraction dataset with 8 predefined schemas (e.g., \textit{"Production Cut"}, where 65\% of test samples contain no matching schemas. Its high schema diversity (3.1 events per sample) challenges multi-schema retrieval.
\item \textbf{SciERC}~\cite{luan-etal-2018-multi}: Cross-domain scientific relation dataset with 15\% schema-free test samples, requiring adaptive extraction of 2.2 relations per sample on average.
\item \textbf{AnatEM}~\cite{DBLP:journals/bioinformatics/PyysaloA14}: Biomedical NER corpus focusing on anatomical entities, featuring strict entity type constraints ideal for closed-schema evaluation.
\item \textbf{ODIE} \cite{jiao-etal-2023-instruct}: Instruction-driven dataset where extraction targets are dynamically specified via natural language.
\end{itemize}

\paragraph{Unified Dataset Construction} 
To simulate real-world multi-domain IE with schema governance, we merge CrudeOilNews, SciERC, and AnatEM into a unified benchmark containing 26 non-overlapping schemas. This setup intentionally avoids schema conflicts to isolate retrieval performance analysis, as current benchmarks lack standardized protocols for overlapping schema evaluation. We downsampled schema-free samples to 30\% of the training set to balance retrieval/generation learning. While limited in scale (unlike ToolenGPT with 200+ tools), this proof-of-concept dataset enables validation of SPT's core mechanisms due to \textit{Schema Independence}. Gradients for each schema token are decoupled due to non-overlapping task allocation, ensuring stable training even with schema scaling. Thus we assume our extraction performance will maintain robustness despite retrieval accuracy may variations.

\subsection{Baselines}
To ensure clarity, we compare our proposed SPT approach to the existing state-of-the-art methods in terms of Schema Retrieval, Infilling, and Generation, respectively.

\paragraph{Retrieval}
We employed retrieval and fine-tuning baselines to evaluate the effectiveness of our proposed approach. For retrieval models, we first write a description for each schema using OpenAI o3-mini-high, then retrieve schemas with a higher similarity score between the query and schema descriptions.

\textbf{BM25}~\cite{bm25} Traditional sparse retrieval using TF-IDF and document length normalization.

\textbf{BGE-M3}~\cite{bgem3} State-of-the-art dense retriever with multilingual/multi-granularity support.

\textbf{BGE-Reranker-Large}~\cite{bgem3} Reranks BGE-M3's top-50 results using cross-attention.

\textbf{Finetuned} Following the SOTA IE sytems (e.g. KnowCoder~\cite{li-etal-2024-knowcoder}, IEPile~\cite{gui-etal-2024-iepile}), we finetune LLMs with LoRA~\cite{hu2022lora} to generate schema names directly from queries.

Retrieval models (BM25, BGE-M3, and BGE-Reranker) use Recall@$k$=5 as the evaluation metric. In contrast, sequence generation-based methods (Fituning and our approach SPT) generate schemas directly, where $k$ corresponds to the number of schemas produced by the LLM.
Note that conventional IE models are excluded in this stage as they lack explicit schema retrieval mechanisms.

\paragraph{Infilling} To compare the performance of our framework in closed IE tasks, we implement several baseline extraction strategies. We evaluate infilling on three datasets—AnatEM, SciERC, and CrudeOilNews, measured by Macro F1 scores.

\textbf{Zero-shot w/ Gold Schemas} LLM extracts with gold schemas.

\textbf{RAG w/ Gold Schemas} LLM augmented with three query-schema-result examples retrieved by BGE-M3, and extract with gold schemas. 

\textbf{Finetuned} LoRA-tuned LLM on the Unified dataset, inference given input queries \textbf{w/ Gold Schemas} and \textbf{w/o Gold Schemas} (provide top-5 schemas retrieved by BGE-M3)

\paragraph{Generation}
We evaluate SPT's schema generation and infilling capabilities on ODIE dataset, measure by soft matching scores (F1) for header evaluation and ROUGE-L F1 scores for content evaluation. Baselines include
\begin{itemize}
\item \textbf{ALPACA/TÜLU/ODIE/GPT-4} \cite{jiao-etal-2023-instruct}: Instruction-tuned models and Commercial LLM from original ODIE paper.
\item \textbf{RAG}: The LLM is augmented with three query-result examples retrieved by BGE-M3.
\item \textbf{Finetuned}: LoRA-tuned language model on ODIE training set.
\end{itemize}

\subsection{Setup}
In our main experiment, we adopt the Qwen2.5-1.5B-Instruct language model as the backbone. The SPT method augments this model with 28 trainable tokens (26 schema tokens plus the \texttt{<Rej>} and \texttt{<Gen>} tokens). Given the model's hidden dimension of 1536, the total number of trainable parameters in SPT amounts to approximately \(28 \times 1536 \approx 43\text{K}\), which is significantly fewer than a typical LoRA with alpha=8 approach that requires tuning on the order of 1.2M parameters. Training is performed on 64 Ascend 910B4 NPUs over 3 epochs on Phases 1 and 2 (Section 3.4) with a learning rate of \(5 \times 10^{-4}\) and 2 additional epochs on Phase 3 with a learning rate of \(5 \times 10^{-5}\). This setup enables efficient and scalable training across our diverse datasets. Detailed examples from SPT and LLM pipelines are included in Appendix \ref{examples}.

\section{Results}

\subsection{Retrieval}

\begin{table}[ht]
\footnotesize
\centering
\begin{tabular}{lccc}
\toprule
\textbf{Models} & \textbf{CrudeOilNews} & \textbf{SciERC} & \textbf{Unified} \\ 
\midrule
bm25               & 0.42               & 0.79          & 0.25         \\ 
bge-m3             & 0.52               & 0.77          & 0.65         \\ 
bge-reranker       & 0.38               & 0.72          & 0.42         \\ 
Finetuned      & 0.46               & 0.83          & 0.61         \\
Ours     & \textbf{0.76}               & \textbf{0.87}          & \textbf{0.82}         \\
\bottomrule
\end{tabular}
\caption{Schema retrieval performance on CrudeOilNews, SciERC, and the "Unified" dataset.}
\label{tab:retrieval}
\vspace{-1\baselineskip}
\end{table}

As shown in Table~\ref{tab:retrieval}, our approach demonstrates superior performance across all three datasets, achieving significant improvements over all retrieval models and the task-specific finetuned LLM. The results bring us two insights.

Firstly, conventional retrieval models (e.g., BM25, BGE-M3) inherently mismatch the schema retrieval task. They rely on semantic similarity between queries and schema descriptions, which may fail to capture contextual associations between queries and scenarios in which schemas are applied. They struggle to bridge the representation gap even with enriched schema descriptions.

Secondly, compared to finetuned LLM that generates full schema names, our method demonstrates architectural advantages of schema token embeddings. The generation process is simplified by encoding structured schemas as compact, task-specific tokens instead of verbose schema names and arguments. This design mitigates error accumulation inherent in autoregressive generation, where LLM's generation of lengthy schema names (e.g., "Organization-Headquarters-Location") increases exposure to decoding errors and semantic drift. Furthermore, SPT achieves superior parameter efficiency with only 43K trainable parameters (vs. LoRA's 1.2M), reducing overfitting risks in low-resource scenarios while maintaining adaptability through optimized token representations.

\subsection{Infilling}
\begin{table*}[ht]
\centering
\footnotesize
\begin{tabular}{lcccccc}
\toprule
\textbf{Models} & \multicolumn{2}{c}{\textbf{AnatEM}} & \textbf{SciERC} & \multicolumn{3}{c}{\textbf{CrudeOilNews}} \\ 
 & Entity&Reject & Relation  & Trigger&Arguments&Reject\\
\midrule
Zero-shot w/ Gold Schemas &0.44&0.58&0.23&0.16&0.15&0.74 \\ 
RAG w/ Gold Schemas &0.71&0.60&0.35&0.33&0.27&\textbf{0.82} \\
Finetuned w/o Gold Schemas &{0.63}&0.56&0.48&{0.46}&{0.35}&0.38 \\
Finetuned w/ Gold Schemas &\textbf{0.83}&-&0.62&\textbf{0.53}&\textbf{0.52}&- \\ 
Ours &0.75&\textbf{0.71}&\textbf{0.64}&0.40&0.32&0.47 \\
\bottomrule
\end{tabular}
\caption{Extraction performance on different datasets}
\label{tab:extraction}
\end{table*}

As shown in Table~\ref{tab:extraction}, both Zero-shot and RAG methods show poor extraction performance ($\le$ 0.35 on SciERC/CrudeOilNews) while overfitting to rejection (0.74-0.82), indicating LLMs' inability to handle complex IE tasks without finetuning. 

While Finetuned w/ Gold Schemas achieves best scores on entity (0.83) and trigger/argument extraction (0.53/0.52), its rejection performance drops significantly (0.38-0.56), demonstrating poor adaptability to schema-free scenarios.

Our method obtains competitive extraction scores (0.75 entity/0.64 relation/0.40 trigger) while substantially improving rejection (0.71-0.47). 
It is worth mentioning that SPT, as an end-to-end approach, achieves better results than the same end-to-end Finetuned w/o Gold Schemas and only slightly inferior to the strong baseline Finetuned w/ Gold Schemas. This balanced performance demonstrates effective integration of schema retrieval, generation, and infilling under a unified framework.

\subsection{Generation}

\begin{table*}[ht]
\centering
\footnotesize
\begin{tabular}{lccc|cccccccc}
\toprule
& \multicolumn{3}{c|}{Header (F1)} & \multicolumn{8}{c}{Content (ROUGE-L)} \\
% \cmidrule(lr){2-4} \cmidrule(lr){5-8}
& \multicolumn{2}{c}{Category}& Overall & \multicolumn{3}{c}{Difficulty} & \multicolumn{2}{c}{Category} & \multicolumn{2}{c}{Source} &Overall \\
& Fixed& Open& & Easy & Medium & Hard & Fixed & Open & Generate & Retrieve &\\
\midrule
ALPACA$^*$          & 0.65 & 0.45 & 0.59 & 0.26 &0.20 &0.22 &0.25 &0.16 &0.30 &0.21 &0.23 \\
TÜLU$^*$            & 0.77 & 0.49 & 0.69 & 0.43 &0.39 &0.38 &0.42 &0.34 &0.45 &0.39 &0.40 \\
ODIE$^*$     & 0.83 & 0.51 & 0.73 & 0.48 &0.45 &0.43 &0.47 &0.41 &0.49 &0.45 &0.45 \\
GPT-4$^*$           & 0.82 & 0.57 & 0.74 & 0.60 &0.55 &0.61 &0.61 &0.51 &0.65 &0.57 &0.59 \\
\midrule
% Zero-shot         & 0    & 0    & 0    & 0    & 0    & 0    & 0    \\
RAG               & 0.32 & 0.24 & 0.28 &0.15&0.10&0.12&0.14&0.13&0.16&0.11&0.14 \\
Finetuned              & 0.76 & 0.53 & 0.71 &0.47&0.38&0.39&0.43&0.37&0.45&0.41&0.42 \\
Ours              & 0.74 & 0.52 & 0.69 &0.43&0.36&0.34&0.39&0.33&0.41&0.38&0.39  \\
\bottomrule
\end{tabular}
\caption{Results on ODIE: Soft matching scores (F1) for header evaluation and ROUGE-L F1 scores for content evaluation. Results with $^*$ are from the ODIE paper.}
\label{tab:ODIE}
\vspace{-1\baselineskip}
\end{table*}

Table~\ref{tab:ODIE} reports our combined ODIE evaluation results, which include both header evaluation (soft matching F1) and content evaluation (ROUGE-L F1) metrics. The header evaluation is split into two categories—Fixed and Open—with an overall F1 score, while the content evaluation is further decomposed into metrics for Difficulty (Easy, Medium, Hard), Category (Fixed, Open), and Source (Generate, Retrieve), along with an overall ROUGE-L score.

Table~\ref{tab:ODIE} reports our combined ODIE evaluation results. e donot report a zeroshot baseline because the difficulty of the task is too high for a 1.5B pretraind model. For header evaluation, our method achieves an overall F1 of 0.69, which is competitive with the LoRA baseline (0.71) and TÜLU$^*$ (0.69).

Regarding content evaluation, our method yields an overall ROUGE-L score of 0.39, with breakdowns of 0.43 (Easy), 0.36 (Medium), and 0.34 (Hard). These scores are slightly lower than those of LoRA (overall 0.42) across the same metrics. Moreover, when examining the category and source components, our method achieves balanced performance (Category: 0.39 Fixed and 0.33 Open; Source: 0.41 Generate and 0.38 Retrieve) compared to LoRA’s corresponding scores.

It is noteworthy that our method has a extreamly low parameter size, the only trainable parameter \texttt{<Gen>} token embedding is trained on a 1.5B model to facilitate on-the-fly schema generation—whereas the all the other baseline, especially from the ODIE paper which leverages LoRA on a larger 7B model. Despite the smaller model size, our approach attains competitive header evaluation and demonstrates balanced performance across all content evaluation metrics. This suggests that embedding a dedicated \texttt{<Gen>} token can effectively reduce the difficulty of schema generation, yielding robust performance even with fewer parameters.

\subsection{Ablation Studies}

\paragraph{Different LLMs}
\begin{table}[ht]
\centering
\footnotesize
\begin{tabular}{lcccc}
\toprule
\textbf{Models} &Retrieval & Trigger&Arguments&Reject\\
\midrule
Qwen1.5B & 0.76 & 0.39&0.34&0.42\\
Qwen7B & 0.84 & 0.49&0.45&0.47\\
Llama3.2  & 0.79 & 0.46&0.41&0.44\\
Phi3.5 & 0.81 & 0.48&0.45&0.48\\
\bottomrule
\end{tabular}
\caption{LLMs performance on CrudeOilNews dataset.}
\label{tab:Size}
\vspace{-1\baselineskip}
\end{table}

Table~\ref{tab:Size} shows the performance of various LLMs on the CrudeOilNews dataset. We compare two variants of the Qwen2.5 series (1.5B and 7B), Llama3.2-3B, and Phi3.5-mini, all with Instruct version. As expected, larger models yield improved performance: Qwen7B outperforms Qwen1.5B in all metrics, demonstrating that stronger LLM capability benefits our extraction task. Notably, Phi3.5-mini, which employs untied input/output embeddings, achieves competitive results compared to tied-embedding model with bigger size i.e. Qwen7B, suggesting that disentangling the input and output embeddings can ease the optimization challenge when tuning only token embeddings—which is crucial for our approach.

\section{Conclusion}
In this paper, we introduced Schema as Parameterized Tools (SPT), which mirrors schemas as callable tools to handle universal IE paradigms through a single adaptive architecture. By reimagining predefined schemas as parameterized tools, SPT enables flexible schema retrieval, infilling, and on-the-fly generation, thereby bridging the gap between closed, open, and on-demand IE tasks. Our experiments across four distinct IE tasks demonstrate that SPT delivers robust schema retrieval and selection performance while achieving extraction accuracy comparable to LoRA baselines and current leading UIE systems with significantly fewer trainable parameters.
The results highlight the potential of SPT as an efficient and adaptable solution for UIE, particularly in resource-constrained settings.

\section{Limitations}
While our proposed framework shows promising results across various IE tasks, it has several limitations that warrant further investigation. First, due to computational resource constraints, our main experiments were primarily conducted on 1.5B models. Although we include preliminary evaluations on larger models (e.g., Qwen7B), a more comprehensive analysis on larger-scale LLMs is needed to assess the scalability and potential performance gains of our approach. Second, since current benchmarks lack protocols for overlapping schema evaluation (as stated in Section 4.1), our evaluation has been restricted to specific datasets such as CrudeOilNews, SciERC, and AnatEM. Additional experiments on more diverse datasets and in different domains are necessary to validate the generalizability of our method. Finally, while our results indicate that models with untied embeddings (e.g., Phi3.5-mini) may offer advantages in optimizing our objective, further exploration is required to understand how different embedding configurations affect performance across various LLM architectures.

% \clearpage
\bibliography{paper,anthology}

\begin{thebibliography}{50}
\providecommand{\natexlab}[1]{#1}

\bibitem[{Ahn(2006)}]{ahn-2006-stages}
David Ahn. 2006.
\newblock \href {https://aclanthology.org/W06-0901/} {The stages of event extraction}.
\newblock In \emph{Proceedings of the Workshop on Annotating and Reasoning about Time and Events}, pages 1--8, Sydney, Australia. Association for Computational Linguistics.

\bibitem[{Banko et~al.(2007)Banko, Cafarella, Soderland, Broadhead, and Etzioni}]{Banko2007OpenIE}
Michele Banko, Michael~J. Cafarella, Stephen Soderland, Matthew Broadhead, and Oren Etzioni. 2007.
\newblock \href {https://api.semanticscholar.org/CorpusID:207169186} {Open information extraction from the web}.
\newblock In \emph{CACM}.

\bibitem[{Cai et~al.(2024)Cai, Wang, Ma, Chen, and Zhou}]{cai2024large}
Tianle Cai, Xuezhi Wang, Tengyu Ma, Xinyun Chen, and Denny Zhou. 2024.
\newblock \href {https://openreview.net/forum?id=qV83K9d5WB} {Large language models as tool makers}.
\newblock In \emph{The Twelfth International Conference on Learning Representations}.

\bibitem[{Chen et~al.(2024)Chen, Xiao, Zhang, Luo, Lian, and Liu}]{bgem3}
Jianlv Chen, Shitao Xiao, Peitian Zhang, Kun Luo, Defu Lian, and Zheng Liu. 2024.
\newblock \href {https://arxiv.org/abs/2402.03216} {Bge m3-embedding: Multi-lingual, multi-functionality, multi-granularity text embeddings through self-knowledge distillation}.
\newblock \emph{Preprint}, arXiv:2402.03216.

\bibitem[{Ding et~al.(2023)Ding, Qin, Yang, Wei, Yang, Su, Hu, Chen, Chan, Chen et~al.}]{ding2023parameter}
Ning Ding, Yujia Qin, Guang Yang, Fuchao Wei, Zonghan Yang, Yusheng Su, Shengding Hu, Yulin Chen, Chi-Min Chan, Weize Chen, et~al. 2023.
\newblock Parameter-efficient fine-tuning of large-scale pre-trained language models.
\newblock \emph{Nature Machine Intelligence}, 5(3):220--235.

\bibitem[{Efeoglu and Paschke(2024)}]{efeoglu2024retrieval}
Sefika Efeoglu and Adrian Paschke. 2024.
\newblock Retrieval-augmented generation-based relation extraction.
\newblock \emph{arXiv preprint arXiv:2404.13397}.

\bibitem[{Fader et~al.(2011)Fader, Soderland, and Etzioni}]{fader-etal-2011-identifying}
Anthony Fader, Stephen Soderland, and Oren Etzioni. 2011.
\newblock \href {https://aclanthology.org/D11-1142/} {Identifying relations for open information extraction}.
\newblock In \emph{Proceedings of the 2011 Conference on Empirical Methods in Natural Language Processing}, pages 1535--1545, Edinburgh, Scotland, UK. Association for Computational Linguistics.

\bibitem[{Gao et~al.(2023)Gao, Xiong, Gao, Jia, Pan, Bi, Dai, Sun, and Wang}]{gao2023retrieval}
Yunfan Gao, Yun Xiong, Xinyu Gao, Kangxiang Jia, Jinliu Pan, Yuxi Bi, Yi~Dai, Jiawei Sun, and Haofen Wang. 2023.
\newblock Retrieval-augmented generation for large language models: A survey.
\newblock \emph{arXiv preprint arXiv:2312.10997}.

\bibitem[{Gui et~al.(2024)Gui, Yuan, Ye, Zhang, Sun, Liang, and Chen}]{gui-etal-2024-iepile}
Honghao Gui, Lin Yuan, Hongbin Ye, Ningyu Zhang, Mengshu Sun, Lei Liang, and Huajun Chen. 2024.
\newblock \href {https://doi.org/10.18653/v1/2024.acl-short.13} {{IEP}ile: Unearthing large scale schema-conditioned information extraction corpus}.
\newblock In \emph{Proceedings of the 62nd Annual Meeting of the Association for Computational Linguistics (Volume 2: Short Papers)}, pages 127--146, Bangkok, Thailand. Association for Computational Linguistics.

\bibitem[{Guo et~al.(2023)Guo, Li, Jin, Liu, Zeng, Liu, Li, Yang, Bai, Guo et~al.}]{guo2023retrieval}
Yucan Guo, Zixuan Li, Xiaolong Jin, Yantao Liu, Yutao Zeng, Wenxuan Liu, Xiang Li, Pan Yang, Long Bai, Jiafeng Guo, et~al. 2023.
\newblock Retrieval-augmented code generation for universal information extraction.
\newblock \emph{arXiv preprint arXiv:2311.02962}.

\bibitem[{Han et~al.(2020)Han, Gao, Lin, Peng, Yang, Xiao, Liu, Li, Zhou, and Sun}]{han-etal-2020-data}
Xu~Han, Tianyu Gao, Yankai Lin, Hao Peng, Yaoliang Yang, Chaojun Xiao, Zhiyuan Liu, Peng Li, Jie Zhou, and Maosong Sun. 2020.
\newblock \href {https://doi.org/10.18653/v1/2020.aacl-main.75} {More data, more relations, more context and more openness: A review and outlook for relation extraction}.
\newblock In \emph{Proceedings of the 1st Conference of the Asia-Pacific Chapter of the Association for Computational Linguistics and the 10th International Joint Conference on Natural Language Processing}, pages 745--758, Suzhou, China. Association for Computational Linguistics.

\bibitem[{Han et~al.(2024)Han, Gao, Liu, Zhang, and Zhang}]{han2024parameter}
Zeyu Han, Chao Gao, Jinyang Liu, Jeff Zhang, and Sai~Qian Zhang. 2024.
\newblock Parameter-efficient fine-tuning for large models: A comprehensive survey.
\newblock \emph{arXiv preprint arXiv:2403.14608}.

\bibitem[{Hao et~al.(2023)Hao, Liu, Wang, and Hu}]{hao2023toolkengpt}
Shibo Hao, Tianyang Liu, Zhen Wang, and Zhiting Hu. 2023.
\newblock \href {https://openreview.net/forum?id=BHXsb69bSx} {Toolken{GPT}: Augmenting frozen language models with massive tools via tool embeddings}.
\newblock In \emph{Thirty-seventh Conference on Neural Information Processing Systems}.

\bibitem[{Hu et~al.(2021)Hu, Shen, Wallis, Allen-Zhu, Li, Wang, Wang, and Chen}]{hu2021lora}
Edward~J Hu, Yelong Shen, Phillip Wallis, Zeyuan Allen-Zhu, Yuanzhi Li, Shean Wang, Lu~Wang, and Weizhu Chen. 2021.
\newblock Lora: Low-rank adaptation of large language models.
\newblock \emph{arXiv preprint arXiv:2106.09685}.

\bibitem[{Hu et~al.(2022)Hu, yelong shen, Wallis, Allen-Zhu, Li, Wang, Wang, and Chen}]{hu2022lora}
Edward~J Hu, yelong shen, Phillip Wallis, Zeyuan Allen-Zhu, Yuanzhi Li, Shean Wang, Lu~Wang, and Weizhu Chen. 2022.
\newblock \href {https://openreview.net/forum?id=nZeVKeeFYf9} {Lo{RA}: Low-rank adaptation of large language models}.
\newblock In \emph{International Conference on Learning Representations}.

\bibitem[{Jiao et~al.(2023)Jiao, Zhong, Li, Zhao, Ouyang, Ji, and Han}]{jiao-etal-2023-instruct}
Yizhu Jiao, Ming Zhong, Sha Li, Ruining Zhao, Siru Ouyang, Heng Ji, and Jiawei Han. 2023.
\newblock \href {https://doi.org/10.18653/v1/2023.emnlp-main.620} {Instruct and extract: Instruction tuning for on-demand information extraction}.
\newblock In \emph{Proceedings of the 2023 Conference on Empirical Methods in Natural Language Processing}, pages 10030--10051, Singapore. Association for Computational Linguistics.

\bibitem[{Lee et~al.(2022)Lee, Soon, Siew, and Sugianto}]{lee-etal-2022-crudeoilnews}
Meisin Lee, Lay-Ki Soon, Eu~Gene Siew, and Ly~Fie Sugianto. 2022.
\newblock \href {https://aclanthology.org/2022.lrec-1.49/} {{C}rude{O}il{N}ews: An annotated crude oil news corpus for event extraction}.
\newblock In \emph{Proceedings of the Thirteenth Language Resources and Evaluation Conference}, pages 465--479, Marseille, France. European Language Resources Association.

\bibitem[{Lester et~al.(2021)Lester, Al-Rfou, and Constant}]{lester2021power}
Brian Lester, Rami Al-Rfou, and Noah Constant. 2021.
\newblock The power of scale for parameter-efficient prompt tuning.
\newblock \emph{arXiv preprint arXiv:2104.08691}.

\bibitem[{Li et~al.(2023)Li, Fang, Yang, Wang, Ye, Zhao, and Zhang}]{li2023evaluating}
Bo~Li, Gexiang Fang, Yang Yang, Quansen Wang, Wei Ye, Wen Zhao, and Shikun Zhang. 2023.
\newblock Evaluating chatgpt's information extraction capabilities: An assessment of performance, explainability, calibration, and faithfulness.
\newblock \emph{arXiv preprint arXiv:2304.11633}.

\bibitem[{Li et~al.(2013)Li, Ji, and Huang}]{li-etal-2013-joint}
Qi~Li, Heng Ji, and Liang Huang. 2013.
\newblock \href {https://aclanthology.org/P13-1008/} {Joint event extraction via structured prediction with global features}.
\newblock In \emph{Proceedings of the 51st Annual Meeting of the Association for Computational Linguistics (Volume 1: Long Papers)}, pages 73--82, Sofia, Bulgaria. Association for Computational Linguistics.

\bibitem[{Li and Liang(2021)}]{li2021prefix}
Xiang~Lisa Li and Percy Liang. 2021.
\newblock Prefix-tuning: Optimizing continuous prompts for generation.
\newblock \emph{arXiv preprint arXiv:2101.00190}.

\bibitem[{Li et~al.(2024)Li, Zeng, Zuo, Ren, Liu, Su, Guo, Liu, Lixiang, Hu, Bai, Li, Liu, Yang, Jin, Guo, and Cheng}]{li-etal-2024-knowcoder}
Zixuan Li, Yutao Zeng, Yuxin Zuo, Weicheng Ren, Wenxuan Liu, Miao Su, Yucan Guo, Yantao Liu, Lixiang Lixiang, Zhilei Hu, Long Bai, Wei Li, Yidan Liu, Pan Yang, Xiaolong Jin, Jiafeng Guo, and Xueqi Cheng. 2024.
\newblock \href {https://doi.org/10.18653/v1/2024.acl-long.475} {{K}now{C}oder: Coding structured knowledge into {LLM}s for universal information extraction}.
\newblock In \emph{Proceedings of the 62nd Annual Meeting of the Association for Computational Linguistics (Volume 1: Long Papers)}, pages 8758--8779, Bangkok, Thailand. Association for Computational Linguistics.

\bibitem[{Liu et~al.(2025)Liu, Zeng, Huang, xinlong hao, Yu, Li, Wang, Gan, Liu, Yu, WANG, Wang, Ning, Hou, Wang, Wu, Xinzhi, Liu, Wang, Tang, Tu, Shang, Jiang, Tang, Lian, Liu, and Chen}]{liu2025toolace}
Weiwen Liu, Xingshan Zeng, Xu~Huang, xinlong hao, Shuai Yu, Dexun Li, Shuai Wang, Weinan Gan, Zhengying Liu, Yuanqing Yu, Zezhong WANG, Yuxian Wang, Wu~Ning, Yutai Hou, Bin Wang, Chuhan Wu, Wang Xinzhi, Yong Liu, Yasheng Wang, Duyu Tang, Dandan Tu, Lifeng Shang, Xin Jiang, Ruiming Tang, Defu Lian, Qun Liu, and Enhong Chen. 2025.
\newblock \href {https://openreview.net/forum?id=8EB8k6DdCU} {Tool{ACE}: Enhancing function calling with accuracy, complexity, and diversity}.
\newblock In \emph{The Thirteenth International Conference on Learning Representations}.

\bibitem[{Lu et~al.(2022)Lu, Liu, Dai, Xiao, Lin, Han, Sun, and Wu}]{lu-etal-2022-unified}
Yaojie Lu, Qing Liu, Dai Dai, Xinyan Xiao, Hongyu Lin, Xianpei Han, Le~Sun, and Hua Wu. 2022.
\newblock \href {https://doi.org/10.18653/v1/2022.acl-long.395} {Unified structure generation for universal information extraction}.
\newblock In \emph{Proceedings of the 60th Annual Meeting of the Association for Computational Linguistics (Volume 1: Long Papers)}, pages 5755--5772, Dublin, Ireland. Association for Computational Linguistics.

\bibitem[{Luan et~al.(2018)Luan, He, Ostendorf, and Hajishirzi}]{luan-etal-2018-multi}
Yi~Luan, Luheng He, Mari Ostendorf, and Hannaneh Hajishirzi. 2018.
\newblock \href {https://doi.org/10.18653/v1/D18-1360} {Multi-task identification of entities, relations, and coreference for scientific knowledge graph construction}.
\newblock In \emph{Proceedings of the 2018 Conference on Empirical Methods in Natural Language Processing}, pages 3219--3232, Brussels, Belgium. Association for Computational Linguistics.

\bibitem[{McClosky et~al.(2011)McClosky, Surdeanu, and Manning}]{mcclosky-etal-2011-event-extraction}
David McClosky, Mihai Surdeanu, and Christopher Manning. 2011.
\newblock \href {https://aclanthology.org/W11-1806/} {Event extraction as dependency parsing for {B}io{NLP} 2011}.
\newblock In \emph{Proceedings of {B}io{NLP} Shared Task 2011 Workshop}, pages 41--45, Portland, Oregon, USA. Association for Computational Linguistics.

\bibitem[{Mintz et~al.(2009)Mintz, Bills, Snow, and Jurafsky}]{mintz-etal-2009-distant}
Mike Mintz, Steven Bills, Rion Snow, and Daniel Jurafsky. 2009.
\newblock \href {https://aclanthology.org/P09-1113/} {Distant supervision for relation extraction without labeled data}.
\newblock In \emph{Proceedings of the Joint Conference of the 47th Annual Meeting of the {ACL} and the 4th International Joint Conference on Natural Language Processing of the {AFNLP}}, pages 1003--1011, Suntec, Singapore. Association for Computational Linguistics.

\bibitem[{Nguyen et~al.(2016)Nguyen, Cho, and Grishman}]{nguyen-etal-2016-joint-event}
Thien~Huu Nguyen, Kyunghyun Cho, and Ralph Grishman. 2016.
\newblock \href {https://doi.org/10.18653/v1/N16-1034} {Joint event extraction via recurrent neural networks}.
\newblock In \emph{Proceedings of the 2016 Conference of the North {A}merican Chapter of the Association for Computational Linguistics: Human Language Technologies}, pages 300--309, San Diego, California. Association for Computational Linguistics.

\bibitem[{Pang et~al.(2023)Pang, Cao, Ding, and Luo}]{pang-etal-2023-guideline}
Chaoxu Pang, Yixuan Cao, Qiang Ding, and Ping Luo. 2023.
\newblock \href {https://doi.org/10.18653/v1/2023.emnlp-main.950} {Guideline learning for in-context information extraction}.
\newblock In \emph{Proceedings of the 2023 Conference on Empirical Methods in Natural Language Processing}, pages 15372--15389, Singapore. Association for Computational Linguistics.

\bibitem[{Pfeiffer et~al.(2020)Pfeiffer, R{\"u}ckl{\'e}, Poth, Kamath, Vuli{\'c}, Ruder, Cho, and Gurevych}]{pfeiffer2020adapterhub}
Jonas Pfeiffer, Andreas R{\"u}ckl{\'e}, Clifton Poth, Aishwarya Kamath, Ivan Vuli{\'c}, Sebastian Ruder, Kyunghyun Cho, and Iryna Gurevych. 2020.
\newblock Adapterhub: A framework for adapting transformers.
\newblock \emph{arXiv preprint arXiv:2007.07779}.

\bibitem[{Pyysalo and Ananiadou(2014)}]{DBLP:journals/bioinformatics/PyysaloA14}
Sampo Pyysalo and Sophia Ananiadou. 2014.
\newblock \href {https://doi.org/10.1093/BIOINFORMATICS/BTT580} {Anatomical entity mention recognition at literature scale}.
\newblock \emph{Bioinform.}, 30(6):868--875.

\bibitem[{Qi et~al.(2024)Qi, Peng, Wang, Xu, Hou, and Li}]{qi-etal-2024-adelie}
Yunjia Qi, Hao Peng, Xiaozhi Wang, Bin Xu, Lei Hou, and Juanzi Li. 2024.
\newblock \href {https://doi.org/10.18653/v1/2024.emnlp-main.419} {{ADELIE}: Aligning large language models on information extraction}.
\newblock In \emph{Proceedings of the 2024 Conference on Empirical Methods in Natural Language Processing}, pages 7371--7387, Miami, Florida, USA. Association for Computational Linguistics.

\bibitem[{Qian et~al.(2024)Qian, Xiong, Liu, and Liu}]{qian-etal-2024-toolink}
Cheng Qian, Chenyan Xiong, Zhenghao Liu, and Zhiyuan Liu. 2024.
\newblock \href {https://doi.org/10.18653/v1/2024.naacl-long.48} {Toolink: Linking toolkit creation and using through chain-of-solving on open-source model}.
\newblock In \emph{Proceedings of the 2024 Conference of the North American Chapter of the Association for Computational Linguistics: Human Language Technologies (Volume 1: Long Papers)}, pages 831--854, Mexico City, Mexico. Association for Computational Linguistics.

\bibitem[{Robertson and Zaragoza(2009)}]{bm25}
Stephen~E. Robertson and Hugo Zaragoza. 2009.
\newblock \href {https://doi.org/10.1561/1500000019} {The probabilistic relevance framework: {BM25} and beyond}.
\newblock \emph{Found. Trends Inf. Retr.}, 3(4):333--389.

\bibitem[{Sainz et~al.(2024)Sainz, Garc{\'\i}a-Ferrero, Agerri, de~Lacalle, Rigau, and Agirre}]{sainz2024gollie}
Oscar Sainz, Iker Garc{\'\i}a-Ferrero, Rodrigo Agerri, Oier~Lopez de~Lacalle, German Rigau, and Eneko Agirre. 2024.
\newblock \href {https://openreview.net/forum?id=Y3wpuxd7u9} {Go{LLIE}: Annotation guidelines improve zero-shot information-extraction}.
\newblock In \emph{The Twelfth International Conference on Learning Representations}.

\bibitem[{Sang and Meulder(2003)}]{TjongKimSang2003IntroductionTT}
E.~Tjong~Kim Sang and Fien~De Meulder. 2003.
\newblock \href {https://api.semanticscholar.org/CorpusID:2470716} {Introduction to the conll-2003 shared task: Language-independent named entity recognition}.
\newblock In \emph{Conference on Computational Natural Language Learning}.

\bibitem[{Schick et~al.(2023)Schick, Dwivedi-Yu, Dess{\`\i}, Raileanu, Lomeli, Hambro, Zettlemoyer, Cancedda, and Scialom}]{schick2023toolformer}
Timo Schick, Jane Dwivedi-Yu, Roberto Dess{\`\i}, Roberta Raileanu, Maria Lomeli, Eric Hambro, Luke Zettlemoyer, Nicola Cancedda, and Thomas Scialom. 2023.
\newblock Toolformer: Language models can teach themselves to use tools.
\newblock \emph{Advances in Neural Information Processing Systems}, 36:68539--68551.

\bibitem[{Shiri et~al.(2024)Shiri, Nguyen, Moghimifar, Yoo, Haffari, and Li}]{shiri2024decompose}
Fatemeh Shiri, Van Nguyen, Farhad Moghimifar, John Yoo, Gholamreza Haffari, and Yuan-Fang Li. 2024.
\newblock Decompose, enrich, and extract! schema-aware event extraction using llms.
\newblock \emph{arXiv preprint arXiv:2406.01045}.

\bibitem[{Stanovsky et~al.(2018)Stanovsky, Michael, Zettlemoyer, and Dagan}]{stanovsky-etal-2018-supervised}
Gabriel Stanovsky, Julian Michael, Luke Zettlemoyer, and Ido Dagan. 2018.
\newblock \href {https://doi.org/10.18653/v1/N18-1081} {Supervised open information extraction}.
\newblock In \emph{Proceedings of the 2018 Conference of the North {A}merican Chapter of the Association for Computational Linguistics: Human Language Technologies, Volume 1 (Long Papers)}, pages 885--895, New Orleans, Louisiana. Association for Computational Linguistics.

\bibitem[{Wang et~al.(2024)Wang, Han, Ji, Wang, Baldwin, and Li}]{wang2024toolgen}
Renxi Wang, Xudong Han, Lei Ji, Shu Wang, Timothy Baldwin, and Haonan Li. 2024.
\newblock Toolgen: Unified tool retrieval and calling via generation.
\newblock \emph{arXiv preprint arXiv:2410.03439}.

\bibitem[{Xiao et~al.(2023)Xiao, Wang, Xu, Wang, Yang, Wang, Luo, Wang, Mao, and Zeng}]{xiao2023yayi}
Xinglin Xiao, Yijie Wang, Nan Xu, Yuqi Wang, Hanxuan Yang, Minzheng Wang, Yin Luo, Lei Wang, Wenji Mao, and Daniel Zeng. 2023.
\newblock Yayi-uie: A chat-enhanced instruction tuning framework for universal information extraction.
\newblock \emph{arXiv preprint arXiv:2312.15548}.

\bibitem[{Xu et~al.(2024{\natexlab{a}})Xu, Chen, Peng, Zhang, Xu, Zhao, Wu, Zheng, Wang, and Chen}]{xu2024large}
Derong Xu, Wei Chen, Wenjun Peng, Chao Zhang, Tong Xu, Xiangyu Zhao, Xian Wu, Yefeng Zheng, Yang Wang, and Enhong Chen. 2024{\natexlab{a}}.
\newblock Large language models for generative information extraction: A survey.
\newblock \emph{Frontiers of Computer Science}, 18(6):186357.

\bibitem[{Xu et~al.(2023)Xu, Xie, Qin, Tao, and Wang}]{xu2023parameter}
Lingling Xu, Haoran Xie, Si-Zhao~Joe Qin, Xiaohui Tao, and Fu~Lee Wang. 2023.
\newblock Parameter-efficient fine-tuning methods for pretrained language models: A critical review and assessment.
\newblock \emph{arXiv preprint arXiv:2312.12148}.

\bibitem[{Xu et~al.(2024{\natexlab{b}})Xu, Li, Xia, and Li}]{xu-etal-2024-enhancing-tool}
Qiancheng Xu, Yongqi Li, Heming Xia, and Wenjie Li. 2024{\natexlab{b}}.
\newblock \href {https://doi.org/10.18653/v1/2024.findings-emnlp.561} {Enhancing tool retrieval with iterative feedback from large language models}.
\newblock In \emph{Findings of the Association for Computational Linguistics: EMNLP 2024}, pages 9609--9619, Miami, Florida, USA. Association for Computational Linguistics.

\bibitem[{Yadav and Bethard(2018)}]{yadav-bethard-2018-survey}
Vikas Yadav and Steven Bethard. 2018.
\newblock \href {https://aclanthology.org/C18-1182/} {A survey on recent advances in named entity recognition from deep learning models}.
\newblock In \emph{Proceedings of the 27th International Conference on Computational Linguistics}, pages 2145--2158, Santa Fe, New Mexico, USA. Association for Computational Linguistics.

\bibitem[{Yakovlev et~al.(2024)Yakovlev, Nikolenko, and Bout}]{yakovlev-etal-2024-toolken}
Konstantin Yakovlev, Sergey Nikolenko, and Andrey Bout. 2024.
\newblock \href {https://doi.org/10.18653/v1/2024.findings-emnlp.345} {Toolken+: Improving {LLM} tool usage with reranking and a reject option}.
\newblock In \emph{Findings of the Association for Computational Linguistics: EMNLP 2024}, pages 5967--5974, Miami, Florida, USA. Association for Computational Linguistics.

\bibitem[{Yuan et~al.(2024)Yuan, Chen, Wang, Fung, Peng, and Ji}]{yuan2024craft}
Lifan Yuan, Yangyi Chen, Xingyao Wang, Yi~Fung, Hao Peng, and Heng Ji. 2024.
\newblock \href {https://openreview.net/forum?id=G0vdDSt9XM} {{CRAFT}: Customizing {LLM}s by creating and retrieving from specialized toolsets}.
\newblock In \emph{The Twelfth International Conference on Learning Representations}.

\bibitem[{Zaken et~al.(2021)Zaken, Ravfogel, and Goldberg}]{zaken2021bitfit}
Elad~Ben Zaken, Shauli Ravfogel, and Yoav Goldberg. 2021.
\newblock Bitfit: Simple parameter-efficient fine-tuning for transformer-based masked language-models.
\newblock \emph{arXiv preprint arXiv:2106.10199}.

\bibitem[{Zheng et~al.(2024)Zheng, Li, Liu, Liu, Luan, and Wang}]{zheng-etal-2024-toolrerank}
Yuanhang Zheng, Peng Li, Wei Liu, Yang Liu, Jian Luan, and Bin Wang. 2024.
\newblock \href {https://aclanthology.org/2024.lrec-main.1413/} {{T}ool{R}erank: Adaptive and hierarchy-aware reranking for tool retrieval}.
\newblock In \emph{Proceedings of the 2024 Joint International Conference on Computational Linguistics, Language Resources and Evaluation (LREC-COLING 2024)}, pages 16263--16273, Torino, Italia. ELRA and ICCL.

\bibitem[{Zhong and Chen(2021)}]{zhong-chen-2021-frustratingly}
Zexuan Zhong and Danqi Chen. 2021.
\newblock \href {https://doi.org/10.18653/v1/2021.naacl-main.5} {A frustratingly easy approach for entity and relation extraction}.
\newblock In \emph{Proceedings of the 2021 Conference of the North American Chapter of the Association for Computational Linguistics: Human Language Technologies}, pages 50--61, Online. Association for Computational Linguistics.

\end{thebibliography}
\clearpage

\appendix

\section{Examples}
\label{examples}
In this appendix we illustrate representative examples from SPT and LLM pipelines, where angle brackets < > indicate one token.

\subsection{SPT}

Figure~\ref{example:spt_retrieval} and Figure~\ref{example:spt_generation} show examples of SPT retrieval and generation modes, respectively.

\begin{figure}[ht]
\footnotesize
\begin{tcolorbox}[
colback=gray!10!white,
colframe=black!75!black,
colbacktitle=black!60!white, % 标题栏背景颜色
title=SPT retrieval mode.,  
width=0.49\textwidth, 
]

[Prompt]

You are a helpful assistant. Select functions to extract structured information. \\

[Input]

Brent dips below \$111, Libya says oil crisis is over. \\

[Output]

Selected function:\\
<MOVEMENT-DOWN-LOSS>, <CRISIS> \\
Extraction results: \\
\{`<MOVEMENT-DOWN-LOSS>': \{`event\_trigger': `dips', `ITEM': `Brent', `FINAL\_VALUE': `below \$111'\}\} \\
\{`<CRISIS>': \{`event\_trigger': `crisis', `ITEM': `oil', `PLACE': `Libya'\}\} \\

\end{tcolorbox}
\caption{SPT retrieval mode.}
\label{example:spt_retrieval}
\end{figure}

\begin{figure}[ht]
\footnotesize
\begin{tcolorbox}[
colback=gray!10!white,
colframe=black!75!black,
colbacktitle=black!60!white, % 标题栏背景颜色
title=SPT generation mode.,  
width=0.49\textwidth, 
]

[Prompt]

You are a helpful assistant. Select functions to extract structured information. \\

[Input]

John, a professor at MIT, founded TechCorp in 2010. \\

[Output]

Selected function: \\
<Rej> \\
Based on the query, I should extract using schema:\\
\{`name': `<Gen>', `arguments': \{`Person': `', `Year': `', `Found Organization': `'\}\}\\
Extraction results: \\
\{`name': `<Gen>', `arguments': \{`Person':`John', `Year': `2010', `Found Organization': `TechCorp'\}\}\\

\end{tcolorbox}
\caption{SPT generation mode.}
\label{example:spt_generation}
\end{figure}

\subsection{LLM}

Figure~\ref{example:llm_retrieval},~\ref{example:llm_extract} and~\ref{example:llm_generate} show examples of LLM retrieval, extraction, and generation, respectively.

\begin{figure}[ht]
\footnotesize
\begin{tcolorbox}[
colback=gray!10!white,
colframe=black!75!black,
colbacktitle=black!60!white, % 标题栏背景颜色
title=LLM retrieval.,  
width=0.49\textwidth, 
]

[Prompt]

You are a helpful assistant. Selecting schemas to extract structured information. \\
Schema List: \\
\{`Event Record': \{`time': `', `location': `', `people': `', 'result': `'\}\},\\
\{`Address Detail': \{`Country': `', `Province': `', `City': `', `Postcode': `'\}\},\\
\{`Company Founding': \{`Founder': `', `Organization': `', `Year': `'\}\}, \\
...\\

[Input]

John, a professor at MIT, founded TechCorp in 2010. \\

[Output]

Company Founding \\

\end{tcolorbox}
\caption{LLM retrieval.}
\label{example:llm_retrieval}
\end{figure}

\begin{figure}[ht]
\footnotesize
\begin{tcolorbox}[
colback=gray!10!white,
colframe=black!75!black,
colbacktitle=black!60!white, % 标题栏背景颜色
title=LLM extraction.,  
width=0.49\textwidth, 
]

[Prompt]

You are a helpful assistant. Extract information using the given, output empty if no matched schema is found. \\
Schema List: \\
\{`Company Founding': \{`Founder': `', `Organization': `', `Year': `'\}\}\\

[Input]

John, a professor at MIT, founded TechCorp in 2010. \\

[Output]

\{`Company Founding': \{`Founder': `John', `Organization': `TechCorp', `Year': `2010'\}\} \\

\end{tcolorbox}
\caption{LLM extraction.}
\label{example:llm_extract}
\end{figure}

\begin{figure}[ht]
\footnotesize
\begin{tcolorbox}[
colback=gray!10!white,
colframe=black!75!black,
colbacktitle=black!60!white, % 标题栏背景颜色
title=LLM generation.,  
width=0.49\textwidth, 
]

[Prompt]

You are a helpful assistant. Generate schemas to extract information. \\

[Input]

John, a professor at MIT, founded TechCorp in 2010. \\

[Output]

Based on the query, I should extract: <Organization> \\
Extraction results: \{`Organization': `TechCorp'\} \\

\end{tcolorbox}
\caption{LLM generation.}
\label{example:llm_generate}
\end{figure}

\section{Time Complexity }

Table~\ref{tab:time_complexity} shows the comparison of time complexity.

\begin{table}[ht]
\footnotesize
\centering
\resizebox{0.49\textwidth}{!}{
\begin{tabular}{lccc}
\toprule
\textbf{Method} & \textbf{Total Time (s)} &	\textbf{Tokens Generated} &	\textbf{Time per Token (ms)} \\
\midrule
SPT & 9.17 & 89 & 103 \\
LLM & 17.61 & 110 & 146 \\
\bottomrule
\end{tabular}}
\caption{Comparison of average inference speed for extraction.}
\label{tab:time_complexity}
\end{table}

\end{document}